\documentclass[a4paper, 10pt]{article}      

\usepackage[left=1.5cm, right=2cm]{geometry}

\usepackage{graphicx} 
\usepackage{caption}
\usepackage{csquotes}
\usepackage{amsmath} 
\usepackage{amssymb}  
\usepackage[table]{xcolor}

\usepackage[backend=bibtex ,natbib=true,giveninits=true,maxnames=25]{biblatex}
\title{\LARGE \bf
Analyzing Multimodal Integration in the Variational Autoencoder from an Information-Theoretic Perspective
}
\definecolor{Motor}{HTML}{c27ba0}
\definecolor{Vision}{HTML}{ffd966}
\definecolor{Touch}{HTML}{93c47d}
\definecolor{Joint}{HTML}{e06666}
\definecolor{Sound}{HTML}{8e7cc3}

\author{Carlotta Langer$^{2}$, Yasmin Kim Georgie$^{1}$, Ilja Porohovoj$^{1}$,  Verena Vanessa Hafner$^{1}$ and Nihat Ay
\thanks{$^{1}$ Yasmin Kim Georgie, Ilja Porohovoj and Verena Vanessa Hafner are with the Department of Computer Science, Humboldt-Universit\"at zu Berlin, Germany 
        {\tt\small \{yasmin.kim.georgie, hafner\}@informatik.hu-berlin.de, i.porohovoj@gmail.com}}%
\thanks{$^{2}$ Carlotta Langer is with the Hamburg University of Technology (TUHH), Institute for Data Science Foundations
        {\tt\small carlotta.langer@tuhh.de}}%
}

\author{Carlotta Langer$^{1}$, Yasmin Kim Georgie$^{2}$, Ilja Porohovoj$^{2,3}$, \\  Verena Vanessa Hafner$^{2}$ and Nihat Ay$^{1,4,5}$
\\
\normalsize{$^{1}$ \, Institute for Data Science Foundations, Hamburg University of Technology, Hamburg, Germany }\\
\normalsize{$^{2}$ \, Department of Computer Science, Humboldt-Universit\"at zu Berlin, Germany}\\
\normalsize{$^{3}$ \, Federal Institute for Materials Research and Testing, Germany}\\
\normalsize{$^{4}$ \, Santa Fe Institute, Santa Fe, USA}\\
\normalsize{$^{5}$ \, Leipzig University, Leipzig, Germany}}

\newcommand{\MeasureTypeOne}{single modality error }
\newcommand{\MeasureTypeTwo}{loss of precision }
\usepackage{xcolor}
\newif\ifdraft
\drafttrue 
\ifdraft
    \newcommand{\TODO}[1]{\textcolor{red}{{[\textbf{TODO}: #1]}}}
    \newcommand{\CL}[1]{\textcolor{violet}{{[\textbf{Carlotta}: #1]}}}
    
    \newcommand{\KIM}[1]{\textcolor{teal}{{[\textbf{Kim}: #1]}}}
    \newcommand{\IP}[1]{\textcolor{orange}{{[\textbf{Ilja}: #1]}}}
    \newcommand{\VVH}[1]{\textcolor{blue}{{[\textbf{Verena}: #1]}}}

\else
    \newcommand{\TODO}[1]{}
    \newcommand{\CL}[1]{}
    \newcommand{\KIM}[1]{}
    \newcommand{\IP}[1]{}
    \newcommand{\VVH}[1]{}
    \newcommand{\NA}[1]{}
    \renewcommand{\sout}[1]{}
    
    \renewcommand{\textcolor}[2]{#2}
\fi

\newcommand\underbracetext[1]{\raisebox{1ex}{\ensuremath{\underbrace{\hphantom{\text{#1}}}_{\text{\normalsize #1}}}}}
\begin{document}

\maketitle
\pagestyle{plain}
\pagenumbering{arabic}

\vspace{1cm}
\begin{abstract}
Human perception is inherently multimodal. We integrate, for instance, visual, proprioceptive and tactile information into one experience. Hence, multimodal learning is of importance for building robotic systems that aim at robustly interacting with the real world. One potential model that has been proposed for multimodal integration is the multimodal variational autoencoder. A variational autoencoder (VAE) consists of two networks, an encoder that maps the data to a stochastic latent space and a decoder that reconstruct this data from an element of this latent space.
The multimodal VAE integrates inputs from different modalities at two points in time in the latent space and can thereby be used as a controller for a robotic agent. Here we use this architecture and introduce information-theoretic measures in order to analyze how important the integration of the different modalities are for the reconstruction of the input data. Therefore we calculate two different types of measures, the first type is called \MeasureTypeOne  and assesses how important the information from a single modality is  for the reconstruction of this modality or all modalities. Secondly, the measures named \MeasureTypeTwo calculate the impact that missing information from only one modality has on the reconstruction of this modality or the whole vector. 
The VAE is trained via the evidence lower bound, which can be written as a sum of two different terms, namely the reconstruction and the latent loss. The impact of the latent loss can be weighted via an additional variable, which has been introduced to combat posterior collapse. Here we train networks with four different weighting schedules and analyze them with respect to their capabilities for multimodal integration.  
\end{abstract}
\vspace{1cm}


\newpage
\section{INTRODUCTION}

Multimodal integration is the combination of different modalities, such as visual, auditory and olfactory. This is crucial for human perception since it increases its precision. 
The integration of the different modalities also plays an important role in the development of a sensorimotor representation of the body, also called a body schema. This has been identified as one important aspect in the development of a human minimal self, see, for instance, \cite{georgie2019interdisciplinary}. In order to develop robots that are able to interact in the real world we would need to equip them with a robust sensorimotor representation of their body. In \cite{Nguyen2021} the authors identify a representation of multimodal sensorimotor contingencies as a critical component of a robotic self.  

There are various approaches to multimodal learning, which we discuss briefly in the next section. In this article we focus on the multimodal variational autoencoder, introduced in \cite{zambelli2020multimodal}. This architecture is well suited for multimodal integration since there the different modalities are first encoded by independent networks and then they are combined in one latent space. In this latent space a truly multimodal representation of the agent's input modalities can be learned. 

In this work we define four different measures with which this integration can be analyzed in detail. This could be used to identify the primary sense of model, detect whether one or multiple modalities are informative for a certain task or guide the learning of a robot to resemble the phases of human development.
During the early stages of human live the importance of the different modalities change, for example, it has been suggested that touch and smell are highly important senses during early infancy, \cite{BIGELOW2020101494, olfactory}. 

Additionally, we use the measures to analyze the integration capacities of different weighting schedules of the latent loss in the evidence lower bound. An introduction to the VAE and the ELBO is given in Section \ref{Sect:ELBO}. We then discuss the architecture of the multimodal VAE and the specific experiment and training model of \cite{zambelli2020multimodal} in Section \ref{sect:Zambelli}. The different information-theoretic measures are introduced in Section \ref{sect:measures} and we present the results in Section \ref{sect:Results}.

\subsection{Related Work}

\subsubsection{Multimodal Computational Models}
There are various approaches to implement multimodal integration. Here we only provide a brief introduction to this field. 

One survey with a thorough taxonomy of the multimodal learning approach is given in \cite{Baltruaitis2017MultimodalML}. The authors there do not focus on sensory modalities, but on natural language, visual signals and vocal signals. They suggest a differentiation between coordinated and joint multimodal representations, where the modalities of joint representations are projected to the same space, while modalities in coordinated models exist in their own spaces and are coordinated through similarities. 

One problem that arises for these model is that they potentially have to deal with very high dimensional data, like the whole visual field. 
Hence, the authors of  \cite{Noda2014} suggest an  approach to multimodal learning using deep neural networks. Their input data is very high dimensional and includes raw RGB images, sound spectra and joint angles and they use multiple encoders and a central hidden layer for the integration. Similarly, the authors of \cite{jVAE} propose a joint VAE. The architecture from \cite{zambelli2020multimodal}, that we use in this work, can be considered in this context. Additionally, the authors of \cite{Meo2021} use a multimodal VAE as part of an active inference controller.

Apart from variational autoencoders also other architectures for deep neural networks have been explored. The authors in \cite{Ngiam2011} present an approach with a shared representation using restricted Boltzmann machines, for instance, and the authors of \cite{schillaci} use a multimodal forward model that is a combination of convolutional networks and a gating system. 

A more exhaustive overview over multimodal deep generative networks can be found in \cite{doi:10.1080/01691864.2022.2035253}, where the authors discuss 22 different models and their properties in detail. They highlight the advantages of a VAE and categorize the different approaches, similar to \cite{Baltruaitis2017MultimodalML}.  


\subsubsection{Information-theoretic Measures}
Information theory originated from Shannon's theory of communication \cite{Shannon}. There he formalizes a communication channel and defines quantities like entropy and mutual information with which the information flow between sender and receiver can be investigated. 
This framework developed into a rich theory with tools to analyze and optimize various types of channels. Here we only want to quickly mention some important information-theoretic measures, that can also be used for guiding the learning and actions of artificial agents. This includes \enquote{empowerment}, which can be seen as a measure of control of an agent \cite{Klyubin}. Another useful measure is \enquote{predictive information}, which is defined as the mutual information between the past and the future of a time series \cite{Bialek}, for instance the sensory input of an agent. Additionally, the authors of  \cite{seitzer2021causal} define an information-theoretic measure for the causal action influence of reinforcement learning agents.

The loss function of the variational autoencoder, the ELBO, is information-theoretic in nature and consists of cross entropy and relative entropy, also called KL-divergence, terms. Hence, if we want to analyze the multimodal integration inside a VAE it is natural to make use of information-theoretic quantities. 

Here in this work we define four different measures that assess the importance of the different modalities by calculating the KL-divergence between representations generated with a varying amount of information about the modalities. A similar approach to quantifying the importance of an information flow, although in the discrete setting, was used, for instance, in \cite{Langer2020,Langer2021, Langer2024}.

\section{MATERIALS AND METHODS}

\subsection{The VAE and the ELBO} \label{Sect:ELBO}

\begin{figure}[h]
\centering
\includegraphics[width = 0.7\textwidth]{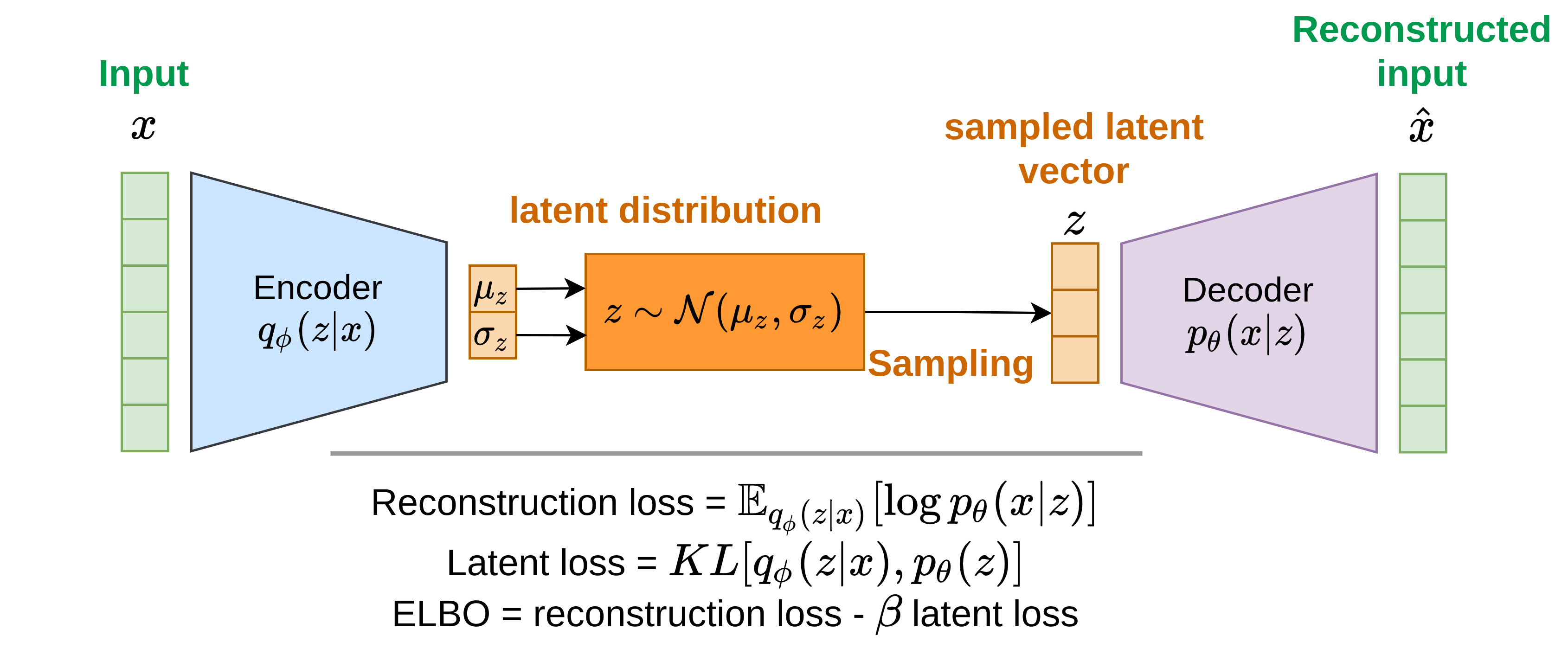}
\captionof{figure}{Architecture of a variational autoencoder with the components of its loss function.} \label{fig:architecture_VAE}
\end{figure}

The variational autoencoder was introduced in \cite{kingma2013auto} and consists of a forward and an inverse model, connected by a latent space, as depicted in Figure \ref{fig:architecture_VAE}. Each input $x$ is mapped to a $\sigma$ and $\mu$ in the latent space, which encodes the Gaussian distribution $q_{\phi}(Z \vert x)$.

The loss function of a VAE is the evidence lower bound, also called ELBO. Figure \ref{fig:architecture_VAE} depicts an illustration of a VAE, consisting of an encoder, a latent space and a decoder. In order to calculate the gradient of the ELBO with respect to $\phi$ one uses the \enquote{reparameterization trick}, as described in \cite{kingma2013auto}. We will now discuss the reasoning of the ELBO in more detail.

\newpage
Let us consider a set of $n$-dimensional data vectors $\{ x_{1}, x_{2}, ... , x_{n} \}$, $x_i \in \mathbb{R}^{n}$. 
Assume that there exists a hidden variable $Z$ that influences the random variable $X$, which generates the samples $x_{i}$. Let $\mathcal{X}, \mathcal{Z}$ be the state spaces of $X$ and $Z$ and let $p_{\theta}$ be a probability density on $\mathcal{X} \times \mathcal{Z}$, parameterized by $\theta$. Then we are able to write the generating process as
\begin{align*}
    p_{\theta}(X,Z) = p_{\theta}(Z)p_{\theta}(X \vert Z).
\end{align*}

In the language of Bayesian Inference the density function $p_{\theta}(Z)$ is called a prior, $p_{\theta}(X \vert Z)$ is the likelihood and  $p_{\theta}(Z \vert X)$ is the posterior. In our setting the posterior is unknown as well as the true parameter $\theta$. As descried in \cite{kingma2013auto} or \cite{Blei2017VariationalInference} in more detail, it is in many cases intractable to optimize the likelihood of the data $p_{\theta}(X)$ or to calculate the true posterior directly, hence the objective is to approximate the posterior by a density  $q_{\phi}$ and to optimize the parameter $\theta$. The prior  $p_{\theta}(Z)$ is fixed, hence independent of $\theta$, to a multivariate standard normal Gaussian distribution with a diagonal covariance matrix in this setting.  

The parameters $\theta$ and $\phi$ are learned using stochastic gradient decent on the following loss function:
\begin{align}
\mathcal{L}(\phi, \theta)  =  \underbrace{\mathbb{E}_{q_{\phi}(z \vert x)} \left[ log \, p_{\theta}(x \vert z) \right]}_{\text{reconstruction error}} - \underbrace{D_{KL}(q_{\phi}(z \vert x) \parallel p_{\theta}(z))}_{\text{latent loss}}. \label{eq:ELBO}
\end{align}
This is called the evidence lower bound, or ELBO, since
\begin{align*}
     log \, p_{\theta}(x) \geq \mathcal{L}(\phi, \theta).
\end{align*}
Hence, by maximizing the ELBO we increase the marginal likelihood, or evidence, of the data.

Now we will take a closer look at the components of the ELBO. The first term in \eqref{eq:ELBO} is called the negative \enquote{reconstruction error}. It takes its maximum value, when $p_{\theta}(x \vert z)  = q_{\phi}(z \vert x)$ for all $z \in \mathcal{Z}$. Therefore it measures how well the VAE can reconstruct given data. The second term in \eqref{eq:ELBO} is called the \enquote{regularizer} or \enquote{latent loss}. Maximizing the ELBO and thereby minimizing the regularizer causes $q_{\phi}(Z \vert x)$ to become similar to the prior $p_{\theta}(Z)$, in our case $\mathcal{N}(0,I)$. This term should prevent the model from over-fitting. The KL-divergence is equal to zero if and only if $q_{\phi}(Z \vert x_i) = p_{\theta}(Z)$ a.e.

In the VAEs applied in \cite{zambelli2020multimodal} the encoder as well as the decoder result in multivariate gaussian densities with diagonal covariance matrices. Let $X$ be an $n$-dimensional gaussian random vector with the mean vector $\mu$ and the covariance matrix $\Sigma$. Then the density function is given by
\begin{align*}
    f_X(x) = \dfrac{1}{\sqrt{ (2 \pi)^n \, \text{det}(\Sigma) }} \exp \left( - \frac{1}{2} (x- \mu)^{\intercal} \Sigma^{-1} (x- \mu) \right),
\end{align*}
where det($\Sigma$) denotes the determinant and $\Sigma^{-1}$ is the inverse. 

Let $\sigma_i^2$ be the $i$th entry on the main diagonal of the covariance matrix. If the covariance matrix is a diagonal matrix, then the density function reduces to
\begin{align}
    f_X(x) = \dfrac{1}{\sqrt{(2 \pi)^n}  \prod\limits_i \sigma_i }      \prod\limits_{i = 1}^n   \exp \left(-\frac{1}{2 \sigma^2_i} (x_i -\mu_i)^2  \right).
\end{align}
Hence the $x_i$ of the random vector $x = (x_1, \dots, x_n)$ are independent of each other, given the latent variable.  

\newpage
\subsubsection{KL-vanishing problem and different weighting schedules}
One common challenge in many VAE models is the KL vanishing problem, also known as \enquote{posterior collapse}. The goal of the VAE is to learn a latent representation with which the data can be reconstructed. 
However, when the posterior of the latent density function \enquote{collapses} into the prior, in our case a standard normal distribution, the generative model of the VAE becomes independent of part of the latent variables, discussed in, for instance, \cite{Alemi2018FixingAB, lucas2019UnderstandingPC}. This will result in reconstructed data that resembles the overall structure of the input data, but is only in a small range and ignores the specific input. 

There are many ways to approach this problem. One way to deal with the posterior collapse is to balance the weights of the two loss terms of the VAE. In that case the KL-divergence corresponding to the latent loss can be weighted by $\beta$, as depicted in Figure \ref{fig:architecture_VAE}. One popular choice of a weighting schedule is to slowly anneal the $\beta$ from 0 to 1, see, for instance, \cite{Bowman2015GeneratingSF, Sonderby2016}. The authors of \cite{fu2019cyclical} propose a cyclical anneal schedule. Setting the $\beta > 1$ leads in \cite{Higgins2016betaVAELB} to a disentangled latent space. 

The authors of \cite{zambelli2020multimodal} state that they focus on the reconstruction and therefore decrease the $\beta$ value from 1 to 0 within the first 1000 steps. Setting $\beta=0$ seems to work well in this case, but it results in a loss function that is not a lower bound for the evidence anymore. Hence, the justification of this strategy requires further theoretical analysis. Here, we want to analyze the behavior of VAE models that follow the structure of \cite{zambelli2020multimodal} and are trained on their data with four different KL-weighting schedules, depicted in Figure \ref{fig:schedules}. The first schedule is constant 1, depicted in the first row. In the second row we show the schedule from \cite{zambelli2020multimodal}, which we call \enquote{constant 0}, since only the first 1000 epochs are trained with a positive $\beta$ value and for the following 79\,000 epochs the $\beta = 0$.

\vspace{-0.25cm}
\begin{center}
    \includegraphics[width = 0.975\textwidth]{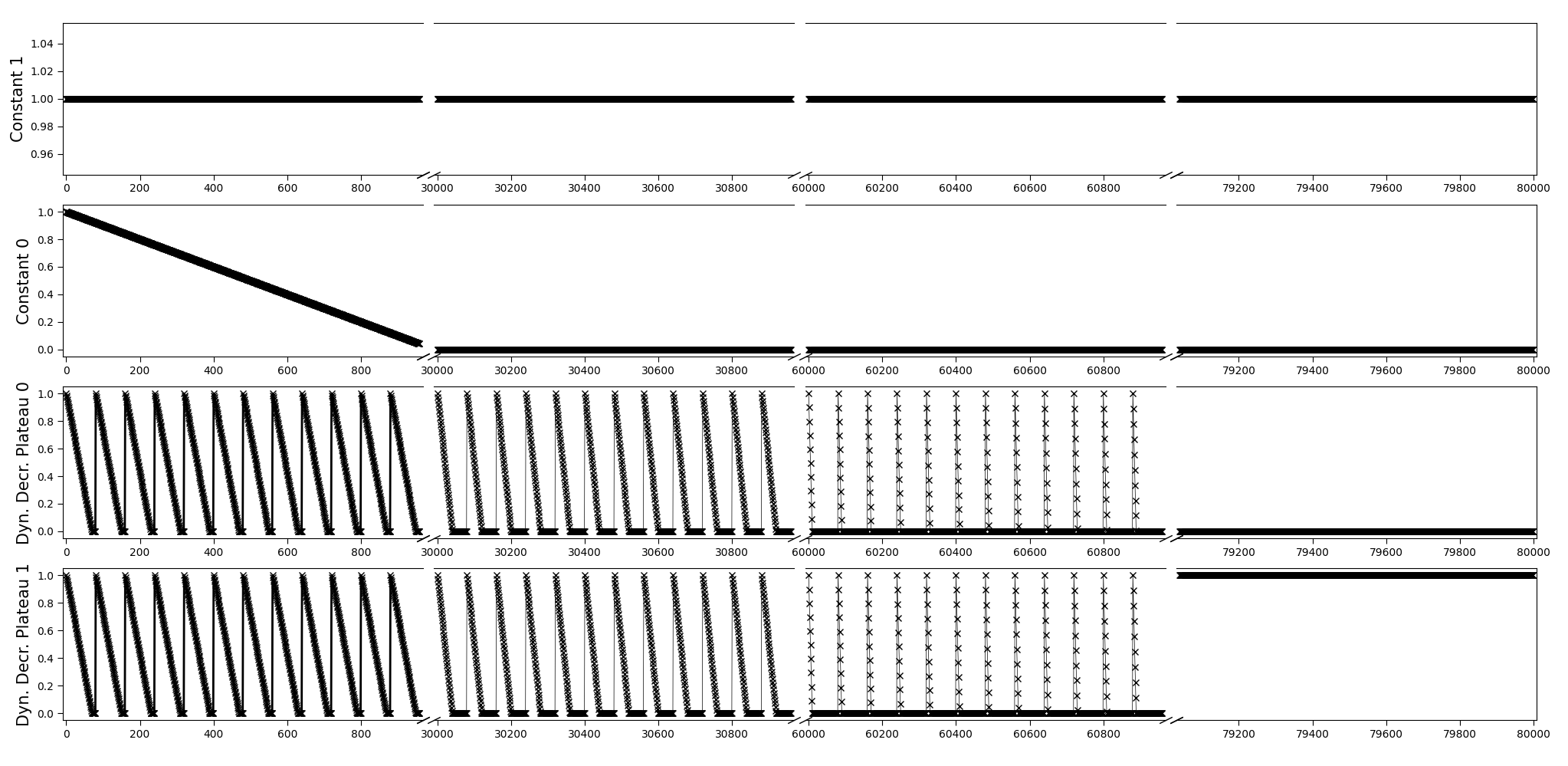}
    \captionof{figure}{ The four different KL-weight schedules, namely \enquote{constant 1}, \enquote{constant 0}, \enquote{dyn. decr. plateau 0} and \enquote{dyn. decr. plateau 1},  depicted for a selected number of epochs.} \label{fig:schedules}
\end{center}

The two schedules in the third and fourth row of Figure \ref{fig:schedules} depict new dynamic schedules. There the $\beta$ decreases from 1 two 0, at first in 80 epochs and then the number of epochs it takes to reach 0 decreases. Since the distance between two epochs with $\beta = 1$ is always 80 epochs there is a \enquote{plateau} at 0 that increases in length during the training. These last two schedules are first equal and only differ in the last 10\,000 epochs, so from 70\,000 to 80\,000. There we set the $\beta$ value of the third schedule to constant 0 and the of fourth schedule to constant 1. In this way we allow the VAE model to converge and we are able to compare the first and the fourth schedule, as well as the second and third, to observe how much the initial dynamic behavior of $\beta$ influences the VAEs.

\subsection{Architecture, Data and Training method} \label{sect:Zambelli}
This section is a brief introduction to the architecture of the multimodal VAE from \cite{zambelli2020multimodal}, their data set and training method. The goal of the multimodal VAE is to integrate different modalities, learn to predict the next sensory input and thereby to control an iCub humanoid robot. This robot receives five different sensory inputs, namely joint position, vision, touch, sound and motor commands. The input vector consists of two time steps for each modality, so for example the joint positions at point $t-1$ and $t$. In this way the model learns a connection between the different points in time.  

The multimodal VAE consists of multiple encoders and decoders for each modality that are then concatenated and integrated into one latent space, see Figure \ref{fig:architecture}. Each encoder and decoder is a neural network, independent of the other modalities. In the reconstruction loss the different modalities are weighted according to their dimensionality to avoid an exaggerated  influence of modalities with more dimensions. Here the latent space that includes all the different modalities has the same dimensionality as the input data, which is 28, since the joint, vision and motor have four dimensions per point in time and  touch and sound one. This seems counterintuitive at first glance, but the reason that the network still learns a meaningful representation and not only copies the input lies in the training approach.

\begin{center}
    \includegraphics[width = 0.5\textwidth]{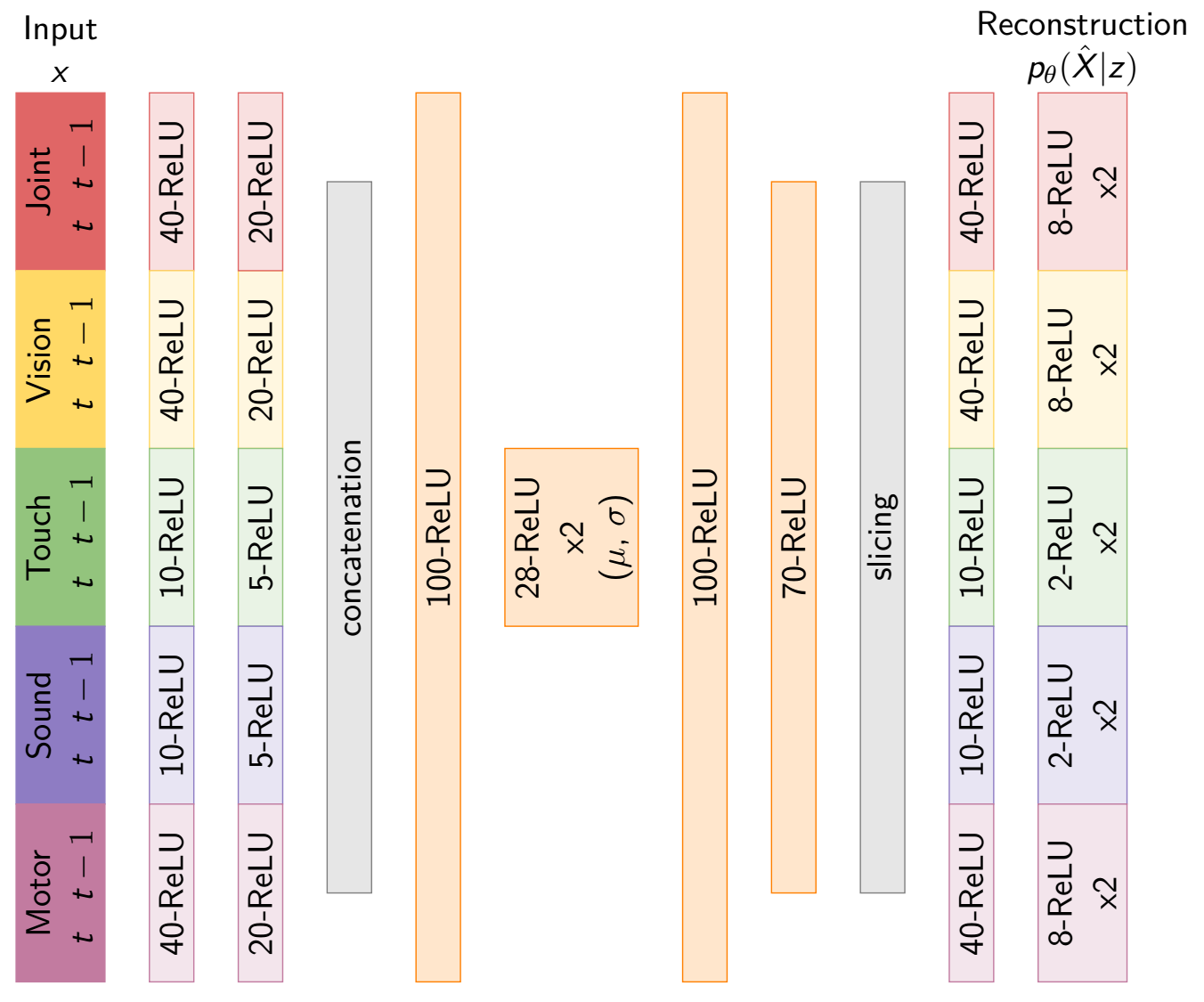}
    \captionof{figure}{Architecture of the multimodal VAE, recreation of Figure 2 of \cite{zambelli2020multimodal}.} \label{fig:architecture}
\end{center}

During training the network not only receives the original dataset, but additionally three types of augmented datapoints. The training data is duplicated and parts of the input is \enquote{muted} by setting its entries to a fixed value out of the range of the input. The input data is normalized between -1 and 1 and the fixed value for the muted instances is -2 in this case. The model then receives the augmented vector as an input and is trained to reconstruct the whole data vector. The types of augmented data points can be seen in Table 1 of \cite{zambelli2020multimodal}.
In this way the model is trained to reconstruct a full data vector from only partial input. When, for instance, all modalities at timepoint $t$ are muted and only the data at timepoint $t-1$ are given, then the model predicts all the sensory and motor states from information only from $t-1$. If part of the input is muted, then it is not possible for the network to simply copy the input data, hence it learns to integrate the different modalities even with a large latent space. More details and learning strategy can be found in Section 3.2 of \cite{zambelli2020multimodal}.

\subsection{Definition of the information-theoretic measures} \label{sect:measures}

In this section we define four different measures for each of the modalities. All of these measures are KL-divergences where we calculate the difference between the reconstruction of a data vector that has information from every modality and the reconstruction where a part of the input $x$ was muted. The method of \enquote{muting} part of the data was briefly described in the previous section and is introduced in more detail in \cite{zambelli2020multimodal}. These density functions are denoted by $p$ and $q$, respectively, in the tables below. The measures differ depending on the part that was muted and whether we calculate the KL-divergence with respect to all of the modalities or only with respect to one modality $M$. 

In each case a set of data points $\{x\}$ is used as an input for the VAE. This set consists of 1024 data points. Let the means of $p(\hat{X} \vert x)$ and $q(\hat{X} \vert x)$ be denoted by the vectors $\mu$ and $\tilde{\mu}$, respectively. Similarly, the diagonal elements of the covariance matrices are denoted by $\sigma_i^2$ and $\tilde{\sigma}_i^2$ . Then we calculate the measures in the following way 
\begin{align}
  \frac{1}{1024} \sum\limits_{x}   D(p(\hat{X} \vert x) \parallel q(\hat{X} \vert x)) &= \frac{1}{1024} \sum\limits_{x} \int\limits_{\hat{x}} p(\hat{x} \vert x) \log \dfrac{p(\hat{x} \vert x)}{q(\hat{x} \vert x)} \notag \\
    &= \frac{1}{1024} \sum\limits_{x}  \frac{1}{2} \left( \sum\limits_{i=1}^n  \log \left( \frac{\tilde{\sigma}_i^2}{{\sigma}_i^2} \right) - n  +  \sum\limits_i \dfrac{\sigma_i^2}{\tilde{\sigma}_i^2}  + \sum\limits_i \dfrac{(\tilde{\mu}_i - \mu_i)^2}{\tilde{\sigma}_i^2}\right). \label{eq:KL}
\end{align}

Let $M$ be one of the modalities. Now we can define the four measures:

\begin{itemize}
\item[] \textbf{Single Modality Error}:

This is the KL-divergence between the reconstruction $p_{\theta}(\hat{X} \vert z)$ given the full input vector $x$ and the reconstruction of a vector where everything is muted \textbf{except for} $x_{M_{t-1}}$.
\begin{itemize}
    \item Regarding only the modality $M$: $\Delta_M(M)$
    
Here the KL-divergence only considers the elements of the input vector $x = (x_1, ... ,x_n)$ that correspond to the modality $M$.
If this measure is close to 0, then $x_{M_t}$ can be reconstructed from $x_{M_{t-1}}$ without using additional information from other modalities.
    \item Regarding all modalities: $\Delta_{all}(M)$
    
 In this case the KL-divergence is calculated with respect to the whole vector $x$. If this measure is close to 0, then all the modalities can be reconstructed from $x_{M_{t-1}}$. In that case $M$ is highly informative for the model, in Section \ref{sect:Results} we will observe that this is the case for the visual modality.
\end{itemize}

\item[] \textbf{Loss of Precision}

For these measures the we calculate the KL-divergence between the reconstruction $p_{\theta}(\hat{X} \vert z)$ given the full input vector $x$ and the reconstruction of a vector where  \textbf{only} $x_{M_{t-1}}$ is muted. 
\begin{itemize}
    \item  Regarding only the modality $M$: $\delta_M(M)$

Here the KL-divergence only considers the elements of the input vector $x$ that correspond to the modality $M$.
If this measure is close to 0, then $x_{M_t}$ can be reconstructed from the other modalities without using additional information from $x_{M_t, M_{t-1}}$.

\item Regarding all modalities: $\delta_{all}(M)$

 In this case the KL-divergence is calculated with respect to the whole vector $x$.
If this measure is close to 0, then $M$ is not essential for the reconstruction of any of the modalities.
\end{itemize}
\end{itemize}

\newpage
The tables below depict which part of the input data was muted for the measures for the modality 
\enquote{Touch}. 

\vspace{0.5cm}
\begin{center}
\begin{minipage}{0.475\textwidth} \footnotesize
    \begin{tabular}{c|p{0.2cm} |p{0.475cm} |p{0.2cm}|p{0.475cm}|p{0.2cm}|p{0.475cm}|p{0.2cm}|p{0.475cm}|p{0.2cm}|p{0.475cm} p{0.001cm}}
   & \multicolumn{2}{c}{Joint \cellcolor{Joint!25} } & \multicolumn{2}{c}{Vision \cellcolor{Vision!25} } & \multicolumn{2}{c}{\cellcolor{Touch!25} Touch} & \multicolumn{2}{c}{\cellcolor{Sound!25} Sound} & \multicolumn{2}{c}{\cellcolor{Motor!25} Motor} \\
 & \centering \cellcolor{Joint!25} $t$ &  \cellcolor{Joint!25} \centering \scriptsize  $t-1$  & \cellcolor{Vision!25} \centering  $t$ &  \cellcolor{Vision!25} \centering \scriptsize  $t-1$  &\centering  \cellcolor{Touch!25} $t$ &\centering \cellcolor{Touch!25} \scriptsize  $t-1$  &\centering \cellcolor{Sound!25} $t$ & \cellcolor{Sound!25} \centering \scriptsize $t-1$  & \cellcolor{Motor!25} \centering  $t$ & \cellcolor{Motor!25}   \scriptsize $t-1$ \\
 \hline 
  p   &  \cellcolor{Joint!25} \centering \checkmark  &  \cellcolor{Joint!25} \centering \checkmark & \cellcolor{Vision!25} \centering \checkmark &   \cellcolor{Vision!25} \centering \checkmark & \cellcolor{Touch!25} \centering \checkmark   & \cellcolor{Touch!25} \centering \checkmark & \cellcolor{Sound!25} \centering \checkmark &  \cellcolor{Sound!25} \centering \checkmark &  \cellcolor{Motor!25} \centering \checkmark &  \cellcolor{Motor!25} \centering \checkmark         &  \\
    q & & & & & & \cellcolor{Touch!25} \centering \checkmark &    \\
       \multicolumn{10}{c}{\hspace{1.4cm} \underbracetext{\textcolor{white}{.....} $\Delta_M$\textcolor{white}{.....}}} \\
            \multicolumn{11}{c}{\hspace{0.2cm}  \underbracetext{\textcolor{white}{...................................} $\Delta_{all}$ \textcolor{white}{...............................} }}
\end{tabular}
\end{minipage} 
\begin{minipage}{0.475\textwidth} \footnotesize
    \begin{tabular}{c|p{0.2cm} |p{0.475cm} |p{0.2cm}|p{0.475cm}|p{0.2cm}|p{0.475cm}|p{0.2cm}|p{0.475cm}|p{0.2cm}|p{0.475cm} p{0.001cm}}
& \multicolumn{2}{c}{Joint \cellcolor{Joint!25} } & \multicolumn{2}{c}{Vision \cellcolor{Vision!25} } & \multicolumn{2}{c}{\cellcolor{Touch!25} Touch} & \multicolumn{2}{c}{\cellcolor{Sound!25} Sound} & \multicolumn{2}{c}{\cellcolor{Motor!25} Motor} \\
 & \centering \cellcolor{Joint!25} $t$ &  \cellcolor{Joint!25} \centering  \scriptsize $t-1$  & \cellcolor{Vision!25} \centering  $t$ &  \cellcolor{Vision!25} \centering  \scriptsize  $t-1$  &\centering  \cellcolor{Touch!25} $t$ &\centering \cellcolor{Touch!25}  \scriptsize  $t-1$  &\centering \cellcolor{Sound!25} $t$ & \cellcolor{Sound!25} \centering  \scriptsize $t-1$  & \cellcolor{Motor!25} \centering  $t$ & \cellcolor{Motor!25}   \scriptsize  $t-1$ \\
 \hline 
    p  &  \cellcolor{Joint!25} \centering \checkmark  &  \cellcolor{Joint!25} \centering \checkmark & \cellcolor{Vision!25} \centering \checkmark &   \cellcolor{Vision!25} \centering \checkmark & \cellcolor{Touch!25} \centering \checkmark   & \cellcolor{Touch!25} \centering \checkmark & \cellcolor{Sound!25} \centering \checkmark &  \cellcolor{Sound!25} \centering \checkmark &  \cellcolor{Motor!25} \centering \checkmark &  \cellcolor{Motor!25} \centering \checkmark         &  \\
    q & \cellcolor{Joint!25} \centering \checkmark  &  \cellcolor{Joint!25} \centering \checkmark & \cellcolor{Vision!25} \centering \checkmark &   \cellcolor{Vision!25} \centering \checkmark &   &  & \cellcolor{Sound!25} \centering \checkmark &  \cellcolor{Sound!25} \centering \checkmark &  \cellcolor{Motor!25} \centering \checkmark &  \cellcolor{Motor!25} \centering \checkmark    &    \\
       \multicolumn{10}{c}{\hspace{1.4cm} \underbracetext{\textcolor{white}{.....} $\delta_M$ \textcolor{white}{.....}}} \\
            \multicolumn{11}{c}{\hspace{0.2cm}  \underbracetext{\textcolor{white}{...................................}  $\delta_{all}$ \textcolor{white}{...............................} }}
            \end{tabular}
\end{minipage}
\captionof{table}{Overview over the parts of the input that is muted for the generation of $p$ and $q$ for the \MeasureTypeOne measures on the left and for the \MeasureTypeTwo measrues on the right. The checkmarks signify that the probabilty density was generated by the VAE with the information of that modality at that time point. If the corresponding cell is empty, then the data was muted there.}
\end{center}

\vspace{0.5cm}
Now we consider the relationship between the measures.
Let $M$ be one of the modalities and let $\hat{X}_M$ denote the random vector of dimension $n_M$ that consists only of the dimensions of $\hat{X}$ that correspond to $M$ in time point $t$ and $t-1$. Let $I= \{1, ... , n\}$ and let $I(M)$ denote the indices of $x_M$ in the whole vector $x$. Following the equation \eqref{eq:KL} we can write 
\begin{align}
    \frac{1}{1024} \sum\limits_{x}  \frac{1}{2} \left( \sum\limits_{i \in I}  \log \left( \frac{\tilde{\sigma}_i^2}{{\sigma}_i^2} \right) - n  +  \sum\limits_{i \in I} \dfrac{\sigma_i^2}{\tilde{\sigma}_i^2}  + \sum\limits_{i \in I} \dfrac{(\tilde{\mu}_i - \mu_i)^2}{\tilde{\sigma}_i^2}\right) =  
    \\   \frac{1}{1024} \sum\limits_{x}  \frac{1}{2} \left( \sum\limits_{i \in I(M)}  \log \left( \frac{\tilde{\sigma}_i^2}{{\sigma}_i^2} \right) - n_M  +  \sum\limits_{i \in I(M)} \dfrac{\sigma_i^2}{\tilde{\sigma}_i^2}  + \sum\limits_{i \in I(M)} \dfrac{(\tilde{\mu}_i - \mu_i)^2}{\tilde{\sigma}_i^2}\right) +  \\   \frac{1}{1024} \sum\limits_{x}  \frac{1}{2} \left( \sum\limits_{i \in I \setminus  I(M)}  \log \left( \frac{\tilde{\sigma}_i^2}{{\sigma}_i^2} \right) - (n-n_M)  +  \sum\limits_{i \in I \setminus I(M)} \dfrac{\sigma_i^2}{\tilde{\sigma}_i^2}  + \sum\limits_{i \in I \setminus I(M)} \dfrac{(\tilde{\mu}_i - \mu_i)^2}{\tilde{\sigma}_i^2}\right)
\end{align}
Since each line line in the equalities above can be expressed as a KL-divergence term and this is non-negative, see Section 8.6 in \cite{Cover}, the following inequalities hold for all modalities $M$: 
\begin{align*}
    \delta_{all}(M) \geq \delta_{M}(M) \\
    \Delta_{all}(M) \geq \Delta_{M}(M).
\end{align*}
The relationship between the \MeasureTypeOne measures and \MeasureTypeTwo measures indicate the influence of a modality. If, for example $\delta_{all}(M) > \Delta_{all}(M)$, then $M$ is highly informative for all other modalities. In that case all modalities can be reconstructed from $M_{t-1}$ with a smaller error compared to a reconstruction where only $M$ is missing. In Figure \ref{fig:ResMultimodal1} we see that this is the case for modality vision and the schedule constant 0.

\section{RESULTS} \label{sect:Results}
In this section we discuss the results of our experiments. For each schedule we trained 20 different models for 80\,000 epochs. The results show the arithmetic mean over the 20 models of one schedule. 

This section is divided into two parts. First, in Section \ref{sect:Results1} we discuss how the different parts of the ELBO behave during the training. There we additionally include the prediction error, which is the euclidean distance between the input $x$ and the mean of the reconstruction $p_{\theta}(\hat{X}\vert z)$, given all the information. Then in Section \ref{sect:Results2} we discuss the different multimodal integration measures. 

\subsection{Loss Functions and Prediction Error} \label{sect:Results1}

Here we discuss the behavior of the different components of the ELBO during the training. In Figure \ref{fig:ELBO_results} we depict the results for the schedules \enquote{constant 1}, \enquote{constant 0}, \enquote{Dyn. Decr. Plateau 0} and  \enquote{Dyn. Decr. Plateau 1} respectively in four columns. The first row shows the prediction error. This is the arithmetic mean over the euclidean distances between all $x$ from a set of 1024 data points and the mean of the reconstructions $\hat{x}$ from the full data vector. In the second row we depict the reconstruction loss, in the third row the latent loss and in the fourth row we have the loss function, the weighted ELBO. Hence, the last row is a combination of the second and third row, where the third row is weighted according to the corresponding $\beta$. 

We observe that the prediction error is the hightest of all the schedules in the case of constant 1, in the first column. Similarly, constant 1 is the schedule with the highest reconstruction loss and the latent loss is close to 0. This is an indication for a posterior collapse, where the posterior distribution collapse into the prior and becomes therefore independent of the input, since the model has a very low latent loss, but a high reconstruction loss and a high prediction error. We will discuss this further in the next section in the context of Figure \ref{fig:Baseline}.

\begin{center}
    \includegraphics[width = \textwidth]{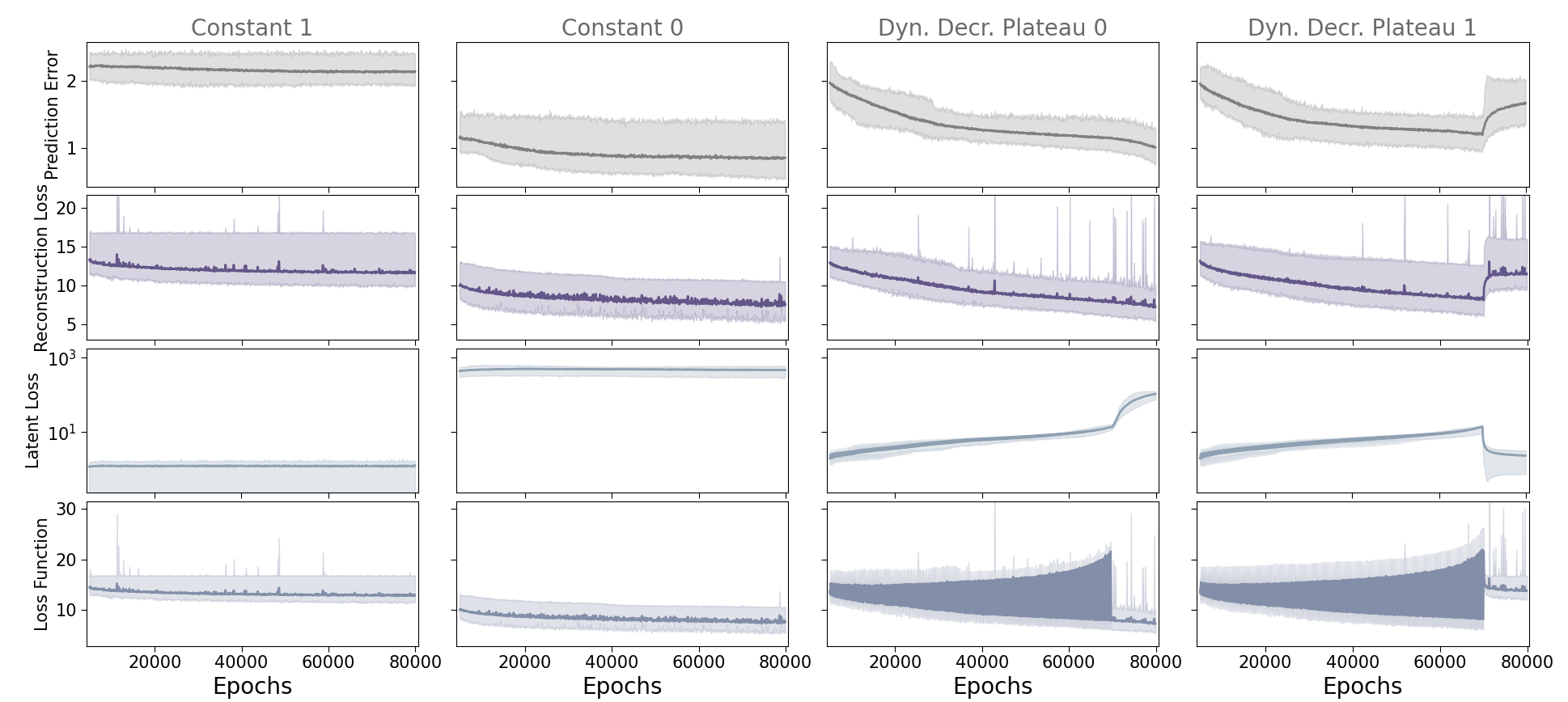}
        \captionof{figure}{Mean, maximum and minimum value of the prediction error, reconstruction loss, latent loss and loss function, each depicted for the four different schedules, namely constant 1, constant 0, dynamic decrease plateau 0 and dynamic decrease plateau 1. } \label{fig:ELBO_results}
\end{center}

The schedule constant 0 does not include the latent loss in its loss function, hence the latent loss is much larger compared to the other schedules. The prediction error and the reconstruction loss are smaller compared to the other schedules, so without any regularizing term the model can fit well to the data. 

The two dynamic schedules have a loss function that changes rapidly for the first 70\,000 epochs, as depicted in the last row. Their prediction error and latent loss lies between the values from the constant schedules, while the reconstruction loss is closer to the value of constant 0. During last 10\,000 epochs the values of the dynamic schedule with a constant value at 0 produces results closer to constant 0 and the dynamic schedule with a constant 1 value has an increase in the loss function and the prediction error, but does not reach the level of the prediction error of constant 1.

\subsection{Measures for the Multimodal Integration} \label{sect:Results2}
Here we discuss the results of the multimodal integration measures. Before we get to the measures introduced in Section \ref{sect:measures} we first consider the values of a measure that we call \enquote{baseline}. This is calculated in the exact same manner as the measures in Section \ref{sect:measures}, but here the whole input vector is muted. In that case the models have no information and generate a full vector with all modalities through sampling from the latent space. We then calculate the KL-divergence between the reconstruction that the model produces when it has all the information and this \enquote{guess}. In Figure \ref{fig:Baseline} we depict these results.  There we observe that constant 1 has a very low baseline. This is another indicator that this schedule suffers from KL-cost vanishing, as the model is not sensitive to the input data. The other schedules have a much higher baseline and we observe that this decreases rapidly for the dynamic schedule in the last column, when the value is set to a constant of one in the last 10\,000 epochs.

\begin{center}
    \includegraphics[width = \textwidth]{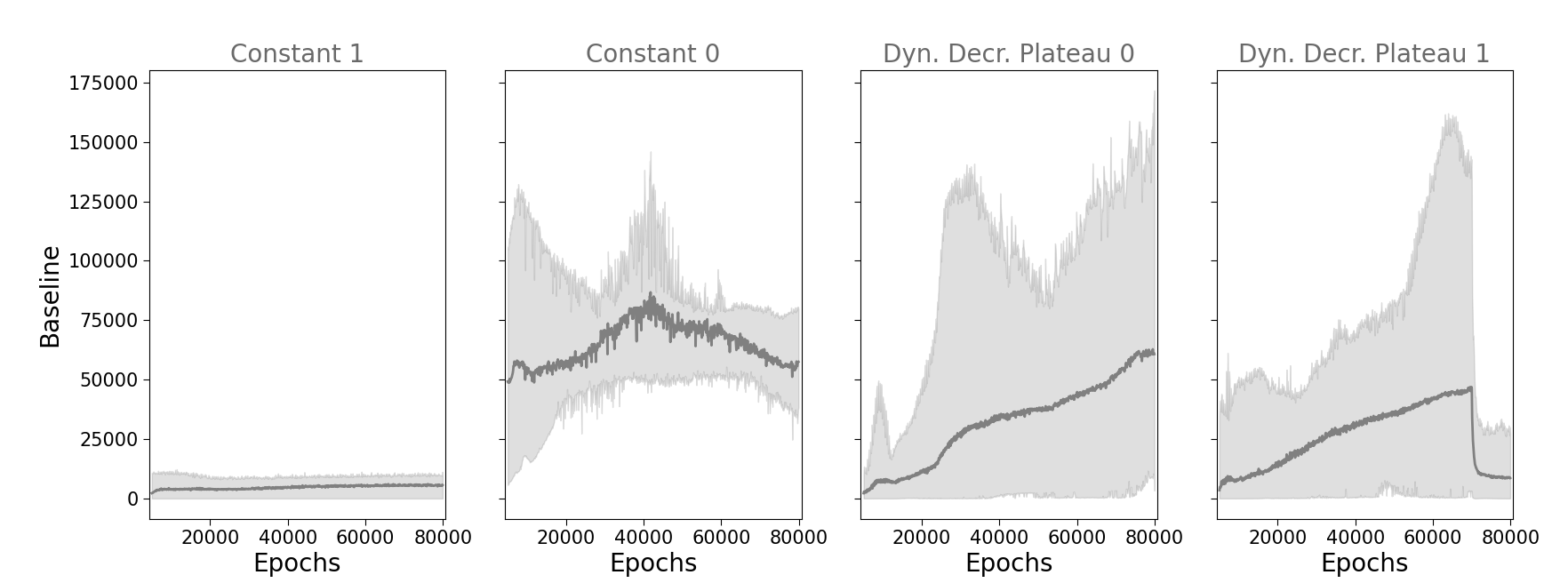}
    \captionof{figure}{Mean, maximum and minimum values of the baseline, KL-divergence between reconstruction from the whole data vector and the reconstruction from a fully muted vector. } \label{fig:Baseline}
\end{center}

Since the baseline is so low for constant 1, we expect that all the other measures are also low, but since the information from the different modalities is not used this does not signify a good multimodal integration. We will therefore tread the results for the constant 1 schedule separately from the others and first discuss the results from the other schedules. In Figure \ref{fig:ResMultimodal1} we see in each column the arithmetic mean over one measure for the different modalities and for the different schedules.

In Figure \ref{fig:ResMultimodal1} in the last three rows we observe that the modality vision has the lowest results in the first two columns and is among the hightest results, especially for constant 0, in the last two columns. This means that vision, as well as the other modalities, can be reconstructed well from vision in $t-1$ alone, as shown by $\Delta_M$ and $\Delta_{all}$, and that the reconstruction gets considerably worse once vision is muted, shown by $\delta_M$ and $\delta_{all}$.

The modalities motor and joint display the opposite dynamic, they are high for $\Delta_M$ and $\Delta_{all}$ and low for $\delta_M$ and $\delta_{all}$. This means that information from these modalities are not useful in reconstructing any of the modalities and that it is possible to reconstruct the joint and motor modalities from the others, as indicated by $\delta_M$ and $\delta_{all}$. The joint positions and motor, in form of velocity commands, are therefore not informative for the other measures alone, but they are integrated in the sense that they can be reconstructed from the other modalities. 

The modalities touch and sound, on the other hand, have overall high results. Especially the observation that  $\delta_M$ and $\delta_{all}$ are very close together indicates that they are hard to predict, regardless of whether the modality in time point $t-1$ is known or not. These are the two modalities with only binary input, either the the robot touches something or hears a sound or not. Hence it might be generally difficult to integrate this binary input.

\begin{center}
    \includegraphics[width = \textwidth]{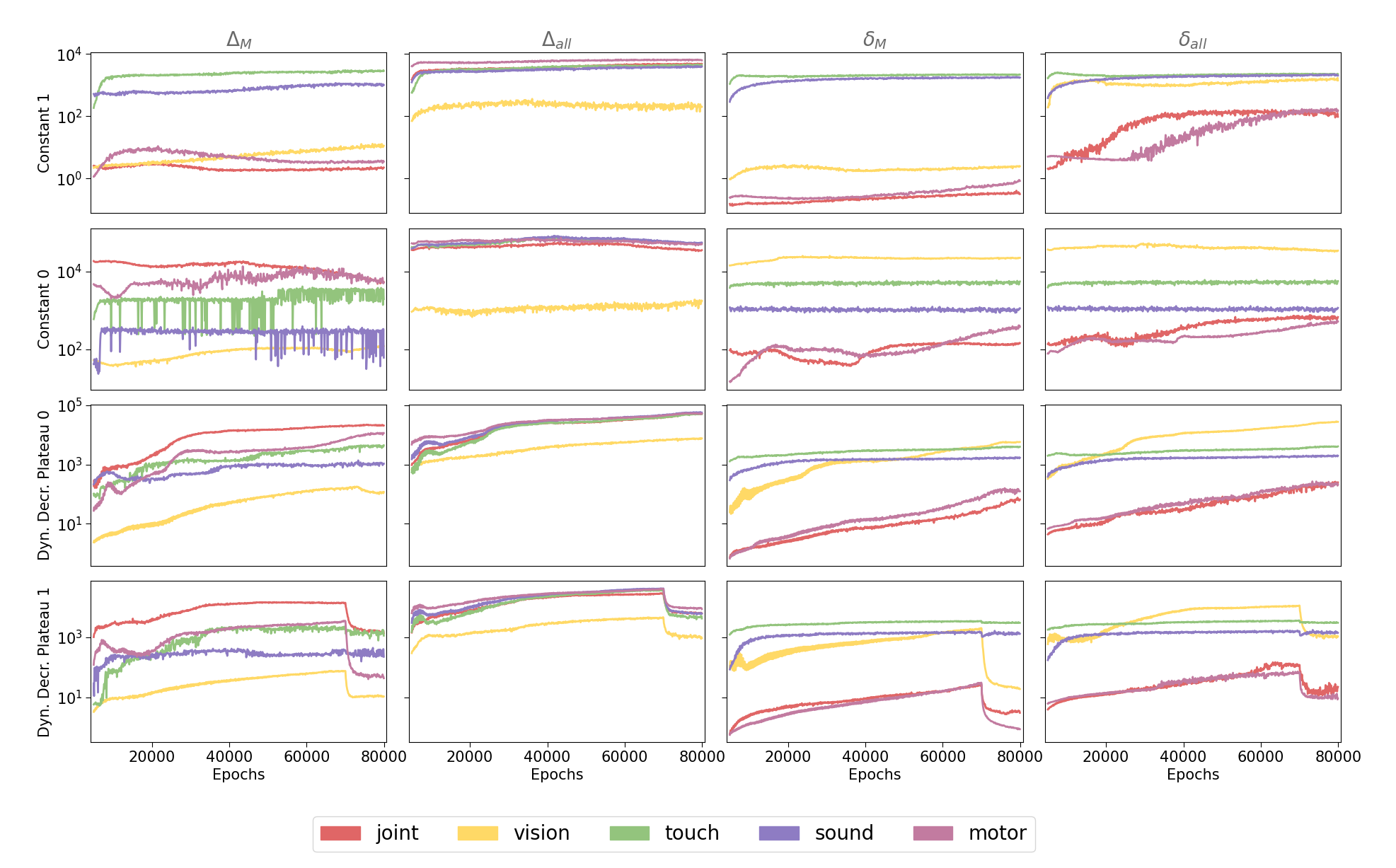}
        \captionof{figure}{The mean results of the four different measures, over 20 training runs each, colored according to the corresponding modality. } \label{fig:ResMultimodal1}
\end{center}

Now we look at the results for the constant 1 schedule. These are lower overall, as expected, since this schedule has a very low baseline. It is therefore not very sensitive to muted data in the input vector. Nonetheless, we also observe similar relationships between the modalities. The results for touch and sound are on a high level and lie very close together, overall. Joint and motor are also low for the measures   $\delta_M$ and $\delta_{all}$, but compared to the other measures they also have a low value for $\Delta_M$. Hence, in this case the model only needs the joint and motor values at time point $t-1$ to predict the results at time point $t$. However, here we do not analyze whether the reconstruction is correct, simply if it changes when some of the input is muted. Since the constant 1 schedule has also a high prediction error the reconstructed distributions for joint and motor might be very different from the actual ones.

Next we compare the behavior of the different measures in Figure \ref{fig:Overview}. The first row depicts the results of the \MeasureTypeOne  measures, and the second one depict the results of the \MeasureTypeTwo measures. The slightly darker marker corresponds to the measure that was calculated with respect to all modalities in each case. In the first column the means of the measures are averaged over the epochs 60\,000-70\,000 and in the second column the means are averaged over the epochs 70\,000-80\,000. Hence, in the first column we consider the results for the dynamic measures before they reach their constant value and in the second column we average over the values once the $\beta$ is fixed to either 0 or 1. 
\begin{center}
        \includegraphics[width = \textwidth]{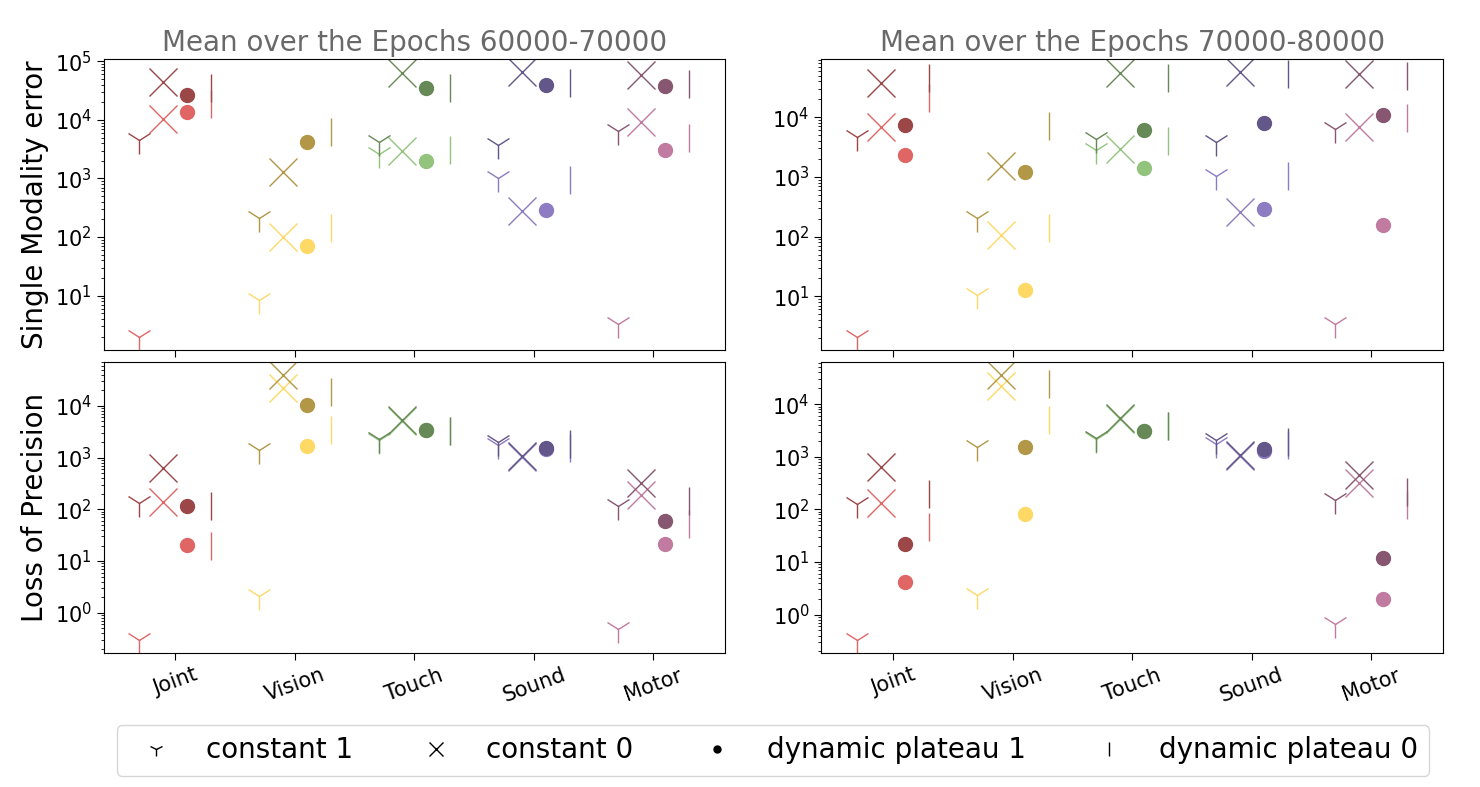}
        \captionof{figure}{Arithmetic mean over the results of the four measures, measure $\Delta_M$ and $\Delta_{all}$ in the first and $\delta_M$ and $\delta_{all}$ in the second row. The mean is taken with respect to 20 models for each schedule and over the epochs 60\,000 to 70\,000 in the first and 70\,000 to 80\,000 in the second column.} \label{fig:Overview}
\end{center}

Here we once again see that the constant 1 schedule hast overall the lowest values. In the left column we observe that the dynamic schedules are very close together. This should be the case, because up to the epoch 70\,000 both schedules are exactly the same. In the second column we see that the dynamic schedule with fixed value 1 has results closer to the constant 1 schedule and the dynamic schedule with a fixed value at 0 moves closer to the results of the constant 0 schedule.  However we can observe that they are not completely equal, so in the case of the plateau at 0 the solution does not completely degenerate to the solution of the constant 0. 
The complete results for all the measures can be found in Appendix \ref{appendix}.
\section{CONCLUSIONS}
In this work we analyze the integration of the different modalities in the multimodal variational autoencoder, introduced in \cite{zambelli2020multimodal}, from an information-theoretic perspective. There we define four different measures that provide insights into the importance of the information from the different modalities. We are able to observe that the visual modality is the most informative one in this setting and that the two binary modalities, namely sound and touch, are very hard to predict overall, regardless of the provided information. 

Furthermore, we analyze models corresponding to four different KL-cost weighting schedules. The model that makes use of the ELBO as originally introduced in \cite{kingma2013auto}, namely constant 1, suffers posterior collapse. Here the reconstruction is not sensitive to the input anymore, hence the measures for multimodal integration are not as informative in this case. The additional calculation of the baseline reveals that the models trained with the constant 1 schedule always reconstruct very similar distributions, regardless whether they have all or no information from the input. 

The constant 0 schedule performs overall best, but since $\beta = 0$ for almost all of the training, these models are not really trained with the ELBO. The loss function only consists of the reconstruction loss and this is not a lower bound for the evidence. Additionally, we analyzed models that were trained with dynamic schedules for the first 70\,000 epochs and then the value of $\beta$ is fixed to either 0 or 1. There we observe that after we fix the value the models become more similar to their constant counterparts, but they do not completely transform into them. Hence, it might be possible to gain a model that maximizes the ELBO while avoiding posterior collapse, by using a posterior schedule at first and then setting $\beta$ to 1.

This work is a first approach to analyzing the multimodal integration in a robot. In future research one might use these measures in order to guide the learning and exploration of a robot. One could use a balance between the different modalities as an intrinsic motivation of such a robot. In this setting we observed, for example, that touch and sound are not very integrated and hard to predict. Hence, a learning robot might need to spend more time observing the hand while it presses on the keyboard. In that way it might be able to learn the connection and thereby to integrate also modalities that are harder to connect to the others. 





\section*{ACKNOWLEDGMENT}
We acknowledge the support of the Deutsche Forschungsgemeinschaft Priority Programme “The Active Self” (SPP 2134).


\printbibliography 

\appendix
\section{Appendix}
\subsection{Complete Results} \label{appendix}
Here are the results for all four measures for every modality and every KL-cost weight schedule. In each of the following figures the first row corresponds to the \MeasureTypeOne measures and the second row depicts the \MeasureTypeTwo measures. As discussed in Section \ref{sect:measures} the inequalties  $ \delta_{all}(M) \geq \delta_{M}(M)$ and $\Delta_{all}(M) \geq \Delta_{M}(M)$ hold. Hence, the upper slightly darker results in the figures below correspond to the measures that were calculated with respect to the whole vector.

\begin{center}
    \includegraphics[width = \textwidth]{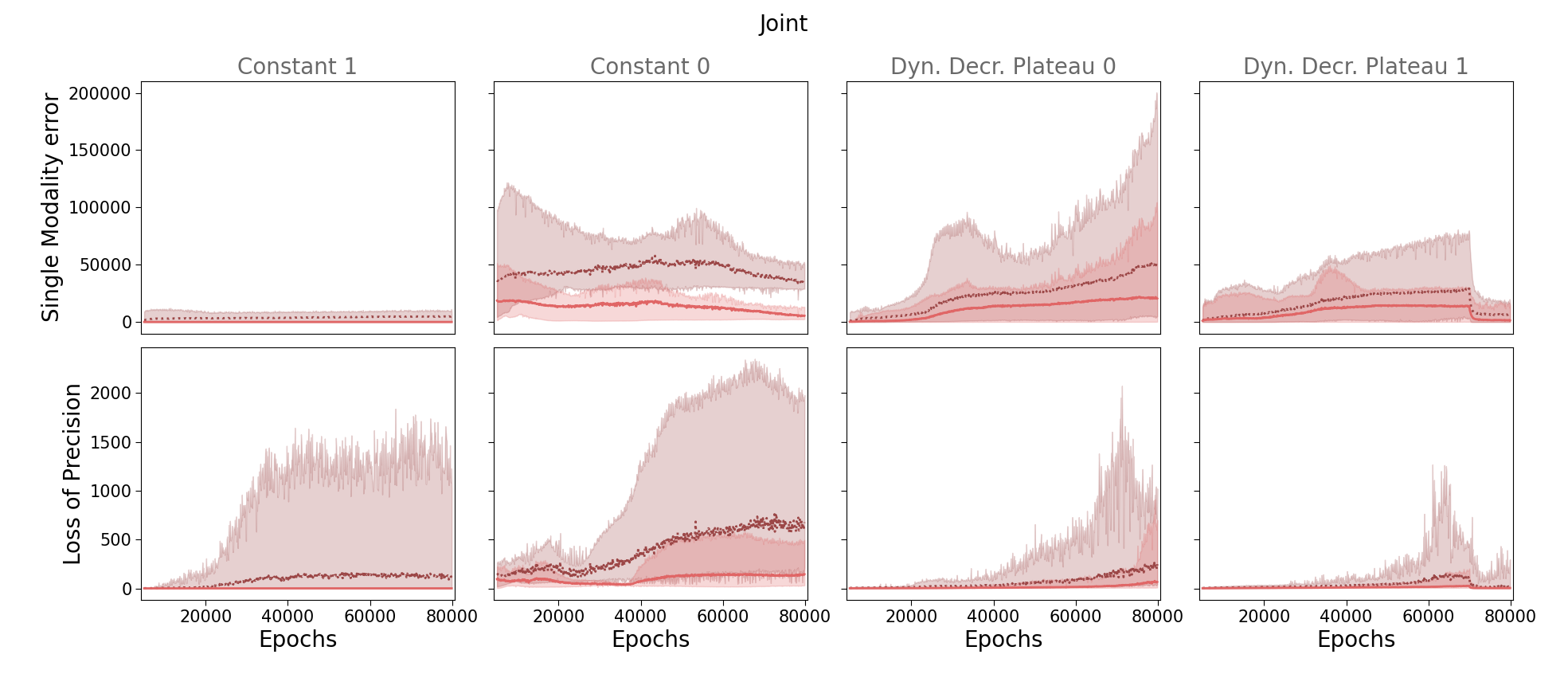}
    \captionof{figure}{ Results for the joint modality with mean, maximum and minimum value.}
\end{center}

\begin{center}
    \includegraphics[width = \textwidth]{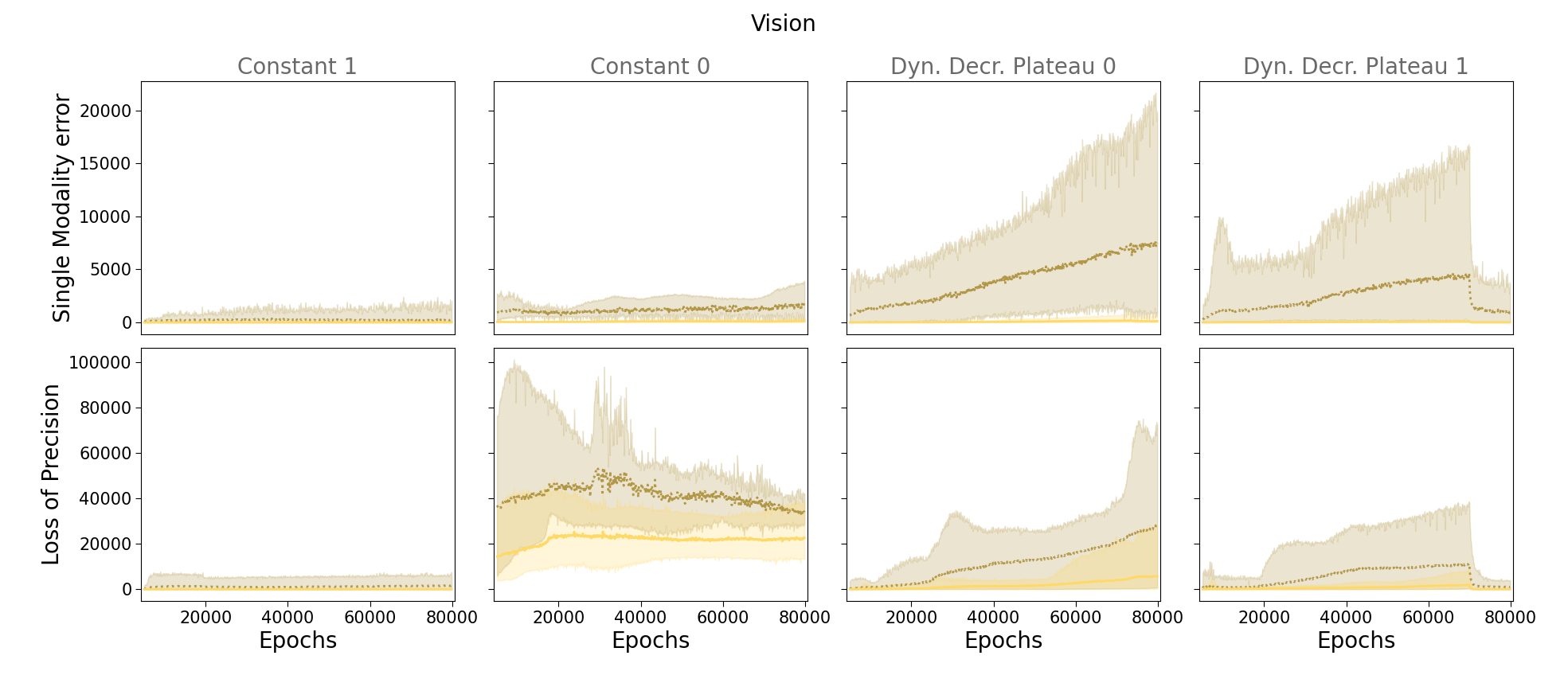}    \captionof{figure}{ Results for the visual modality with mean, maximum and minimum value.}
\end{center}

\begin{center}
    \includegraphics[width = \textwidth]{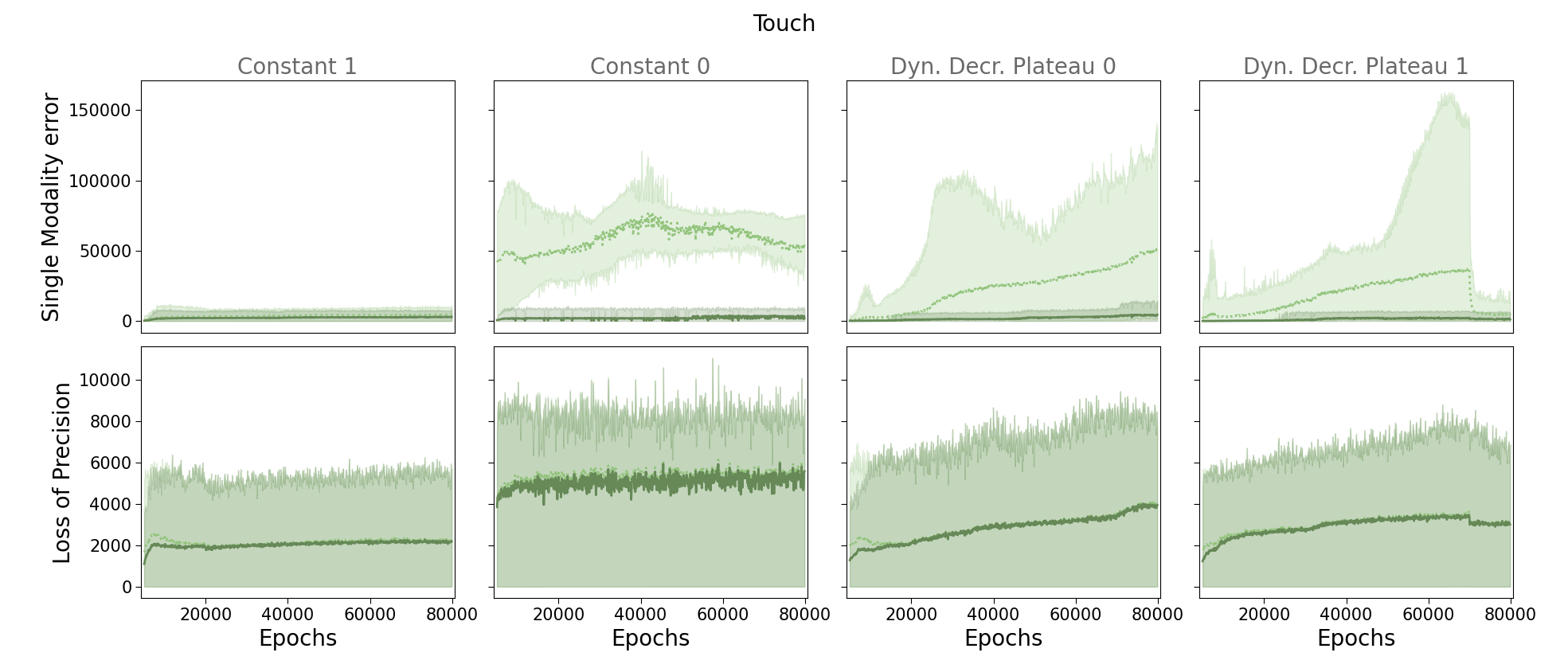}
        \captionof{figure}{ Results for the touch modality with mean, maximum and minimum value.}
\end{center}

\begin{center}
    \includegraphics[width = \textwidth]{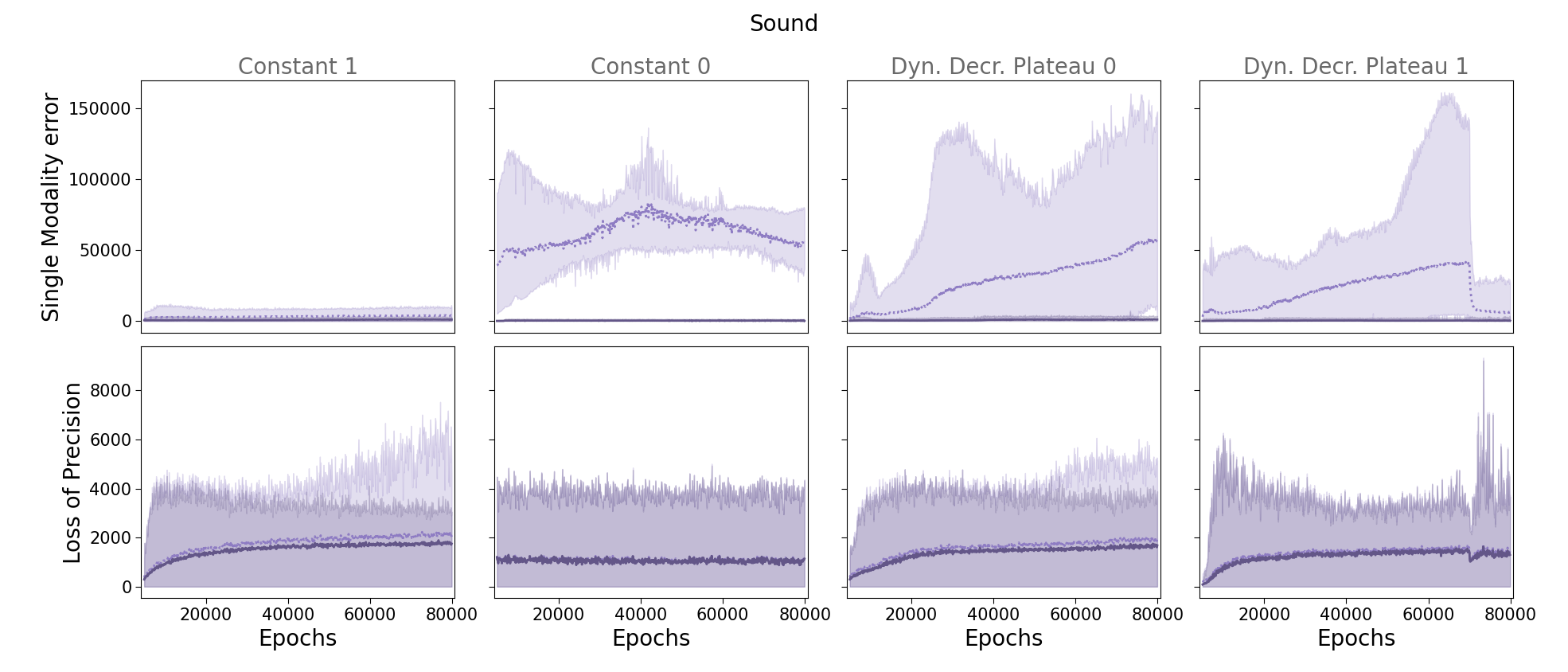}
        \captionof{figure}{ Results for the sound modality with mean, maximum and minimum value.}
\end{center}

\begin{center}
    \includegraphics[width = \textwidth]{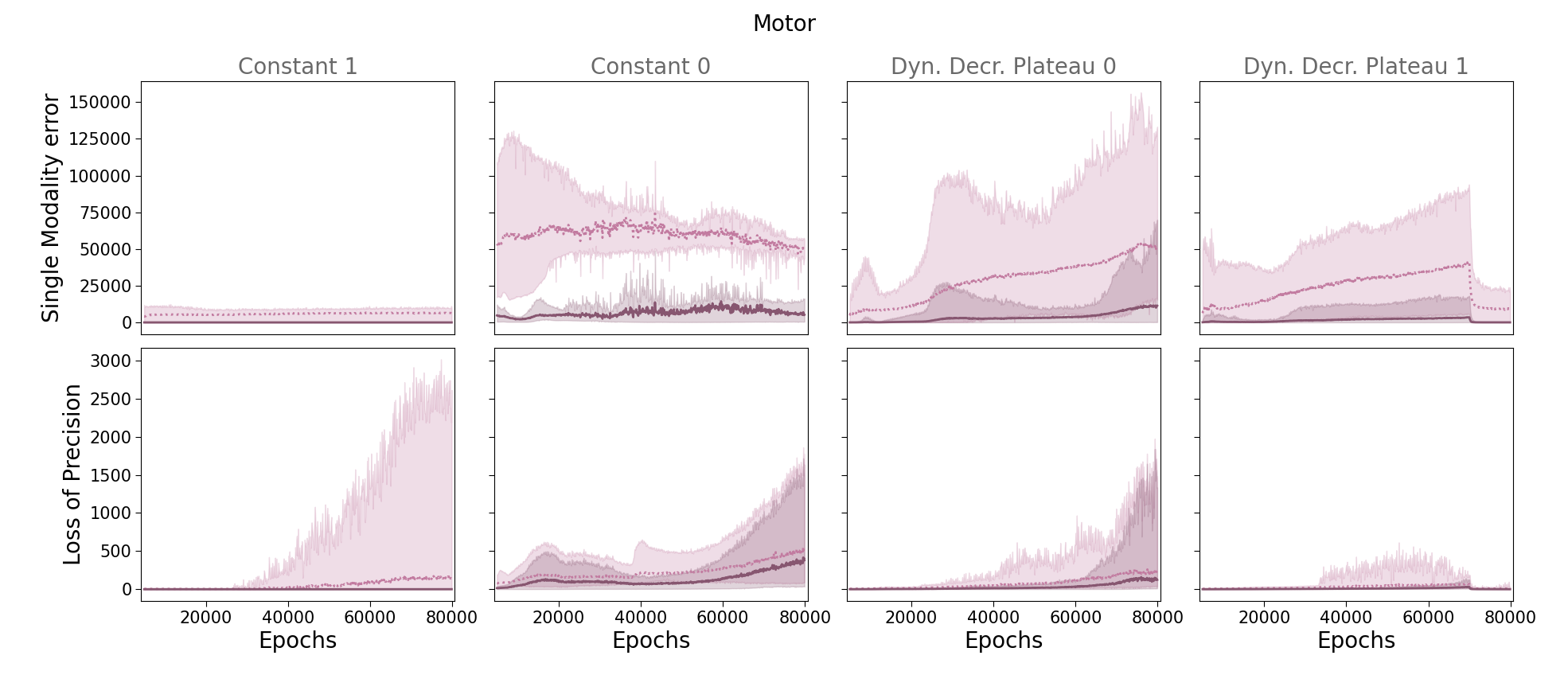}
        \captionof{figure}{ Results for the motor modality with mean, maximum and minimum value.}
\end{center}
\end{document}